\documentclass[conference]{IEEEtran}
\IEEEoverridecommandlockouts
\usepackage{cite}
\usepackage{amsmath,amssymb,amsfonts}
\usepackage{algorithm}
\usepackage{algorithmic}
\usepackage{graphicx}
\usepackage{textcomp}
\usepackage{xcolor}

\def\BibTeX{{\rm B\kern-.05em{\sc i\kern-.025em b}\kern-.08em
    T\kern-.1667em\lower.7ex\hbox{E}\kern-.125emX}}
\begin{document}

\title{Shared MF: A privacy-preserving recommendation system
}

\author{\IEEEauthorblockN{Senci Ying}
\IEEEauthorblockA{\textit{College of Computer Science,} \\
\textit{Zhejiang University}\\
HangZhou,China \\
scying@zju.edu.cn}
}

\maketitle

\begin{abstract}
Matrix factorization is one of the most commonly used technologies in recommendation system. 
With the promotion of recommendation system in e-commerce shopping, online video and other aspects, distributed recommendation system has been widely promoted,
 and the privacy problem of multi-source data becomes more and more important.
 Based on Federated learning technology, this paper proposes a shared matrix factorization scheme called SharedMF.
 Firstly, a distributed recommendation system is built, and then secret sharing technology is used to protect the privacy of local data. 
 Experimental results show that compared with the existing homomorphic encryption methods, our method can have faster execution speed without privacy disclosure, and can better adapt to recommendation scenarios with large amount of data.
\end{abstract}


\section{Introduction}

In order to adapt to the development of the Internet, recommendation system is widely used in e-commerce platform and entertainment application. 
The data scale of these platform is also expanding with the use of users and the items listed in them, more worse the computing pressure of the traditional centralized recommendation system increases, and the performance requirements of hardware are gradually improved. 
On this basis, the distributed recommendation system is gradually rising. Its main idea is to store the data in each node separately, such as the personal data of each user stored on personal devices, and the recommendation system can be implemented through joint modeling. 
At the same time, the conflict between the use of big data and data privacy protection inevitably breaks out. The EU's general data protection regulation (GDPR) is a new challenge to the traditional data processing mode of artificial intelligence.

To solve the privacy protection problem in the distributed model training, federal learning technology was proposed by Google in 2016\cite{konevcny2016federated}.
The framework assumes that the data parties involved in the model training are not trustworthy, and the data can not be exposed to any party. 
Through a series of cryptographic proofs, the model can still complete the joint training with high performance and accuracy without privacy disclosure. 
In the context of distributed recommendation system, it is very important to use federated learning technology to protect privacy. 
We must ensure that information such as user's age, property status and so on cannot be obtained by others in data use. 
Matrix factorization is the most commonly used recommendation system model in industry. It has high computational cost and high accuracy, and is easy to be applied in distributed recommendation environment. 

\begin{figure}[htbp]
    \centerline{\includegraphics[width=0.45\textwidth]{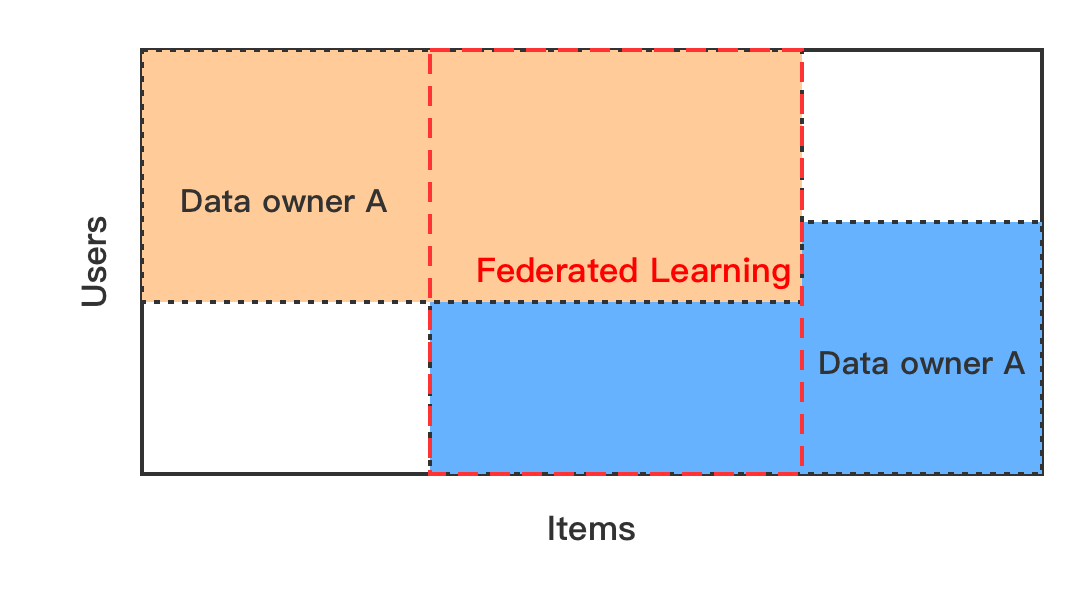}}
    \caption{Federated learning in recommendation system}
    \label{fedml}
\end{figure}

In this paper, a distributed matrix factorization recommendation system based on secret sharing is proposed. We assume that user data is distributed across different data sources, such as Taobao and Amazon, which have different customers. 
Each data sources will calculate its own user vectors locally, and calculate the item vectors in the central server. The problem of privacy disclosure while updating the parameters can be solved by secret sharing. 
The experimental results show that our algorithm has higher efficiency while preserving privacy. Our main contributions are as follows:
\begin{itemize}
    \item We propose a privacy protection recommendation system model, which can not only complete the distributed recommendation task, but avoid privacy disclosure.
    \item Our model is based on secret sharing technology. Compared with the existing schemes based on homomorphic encryption, the running speed of our model is greatly improved.
    \item Our proposal has been verified on real data sets, which fully proves its superiority.
\end{itemize}
\section{Related Work}
In this section, We first review the privacy protection techniques commonly used in federated learning, and then discuss some of their existing applications in recommendation systems. 
\subsection{Federated learning}
The idea of federated learning is to complete the joint modeling by combining the partial data shared by each data source. Its application in the recommendation system is shown in the figure\ref{fedml}. 
The two recommendation platforms have some of the same products, but the customers are not the same. Therefore, the performance of the model can be improved by combining different user data. 
In this process, privacy protection technology is needed to ensure security. Common methods are as follows.
\paragraph{Anonymization}
Anonymization is to hide or generalize the part of data features with privacy exposure. Among them, k-anonymity \cite{samarati1998generalizing} is the most widely used privacy protection method for anonymous data in the data publishing scenario. 
It was proposed by samarati et al. In 1998, it specifies the maximum acceptable information disclosure degree through the parameter K, and requires at least k on the recognizable data field The data is relatively fuzzy, so the attacker can't lock a specific user by obtaining the information of this field.
\paragraph{Homomorphic encryption}
Homomorphic encryption\cite{rivest1978data} can ensure that the results obtained by decrypting the encrypted data directly are consistent with the results of the same operation on the plaintext data. 
In the process of model learning parameter exchange, homomorphic encryption can protect the privacy of user data. The Encryption ensures that the data and the model itself will not be transmitted, thus reducing the privacy leakage at the data level.
\paragraph{Differential privacy}
Differential privacy\cite{dwork2008differential} protection is the problem of privacy leakage caused by small changes in the data source. 
It is difficult for the observer to detect the subtle changes in the data set by observing the output results, so as to achieve the purpose of privacy protection. The common method is to add random noise to the input or output, such as Laplacian noise, in order to cover up the real data
\subsection{Privacy-preserving recommender system}
The related technologies mentioned above are applied in privacy protection recommendation system. in \cite{wei2018improving} the author used a k-anonymity  method in the collaborative filtering and used a potential factor model to reduce the sparsity of the matrix, which protects the data privacy.
There are many privacy protection strategies for recommendation models. Next, we will focus on the privacy protection algorithms related to the matrix factorization model in this paper.

Raghavendran Balu and Teddy Furon\cite{balu2016differentially} use sketching techniques to implicitly provide the differential privacy while training matrix factorization model.Its model scales well with data and is suitable for large scale application.
Sungwook Kim et al\cite{kim2016efficient} design a efficient data structure to use fully homomorphic encryption in matrix factorization, its inputs and outputs are both encrypted so there is no privacy problem while training recommend system.
Qiang Tang and Husen Wang\cite{tang2017privacy} proposed a hybrid recommender model by using incremental matrix factorization which makes the system privacy-preserving and more efficient than other model.
Nonnegative matrix factorization (NMF) has been successfully applied and in\cite{qian2020fast} the author proposed a distributed sketched alternating nonnegative least squares (DSANLS) framework for NMF and show the framework can be adapted to the security setting.
Secure matrix factorization\cite{chai2019secure} used a user-level distributed framework which makes user update the gradient locally and  uploads the encrypted gradient to tune parameters in server.
\begin{figure}[htbp]
    \centerline{\includegraphics[width=0.45\textwidth]{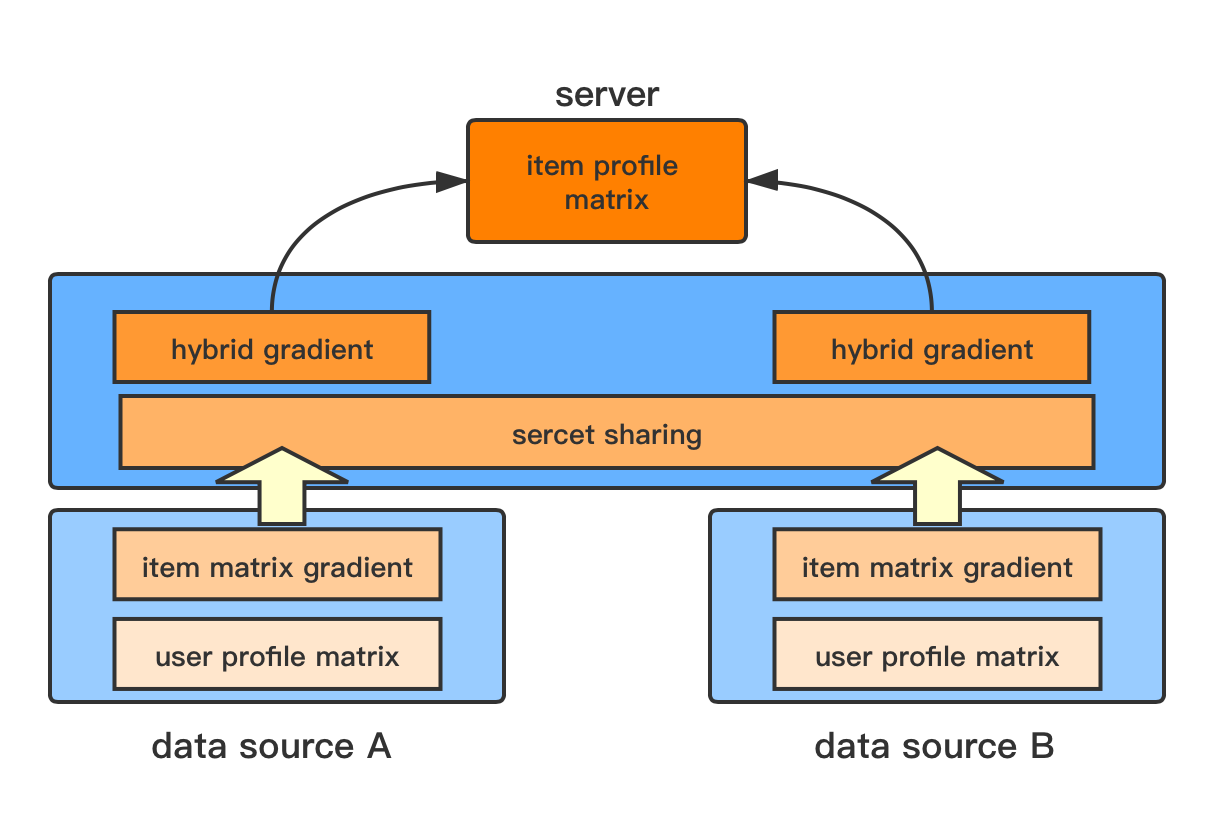}}
    \caption{Shared MF: two data sources example}
    \label{smf}
\end{figure}
\section{Matrix factorization with secret sharing}
Firstly, we propose the background knowledge of matrix factorization. Then, according to the characteristics of multi-source data, we design a distributed matrix factorization recommendation system model, and introduce secret sharing technology to solve the privacy protection problem. 
We name this model as shared matrix factorization and figure\ref{smf} shows its structure. Compared with the existing solutions based on homomorphic encryption, our model's privacy protection strategy will be better for its high efficiency and practicability.
\subsection{Matrix factorization}
Suppose we have n users, m items and each user rated a subset of m items. Then we can get a n*m rating matrix M. let $r_{ij}\in M$ denote the score of item j rating by user i. Because the items have a large amount of numbers, the rating matrix is inevitably sparse, and what we want to do is find a solution to get the unknow rate in M.
Matrix factorization formulates this problem as fitting a bilinear model on the rating matrix. It believes that users and items each have a profile matrix, and the rating matrix is the product of them. In particular, denoting user profile matrix as $U\in R^{n*d}$ and item profile matrix as $V\in R^{d*m}$.
Then the rating $r_{ij}$ can be computed by $\left<u_i,v_j\right>$ where $u_i$ denotes the $i$th row of U and $v_j$ denotes the $j$th column of V.
The solution of U,V can be found by solving the following equation \eqref{lsm}:
\begin{equation}
    \min_{U,V}=\frac{1}{M}\sum_{i=1}^{n}\sum_{j=1}^{m}(r_{ij}-\left<u_i,v_j\right>)^2+\lambda||U||^2_2+\mu||V||^2_2
    \label{lsm}
\end{equation}

where $\lambda$ and $\mu$ are the regularization parameters. And we can solve the above problem by stochastic gradient descent with the following equations\cite{koren2009matrix}:
\begin{equation}
    u_i^{new} = u_i^{old}-\alpha\Delta_{u_i}F(U^{old},V^{old})
    \label{sgd1}
\end{equation}
\begin{equation}
    v_j^{new} = v_j^{old}-\alpha\Delta_{v_j}F(U^{old},V^{old})
    \label{sgd2}
\end{equation}
where
\begin{equation}
    \Delta_{u_i}F(U,V) = -2\sum_{j=1}^{m}v_j(r_{ij}-\left<u_i,v_j\right>)+2\lambda u_i
    \label{sgd3}
\end{equation}
\begin{equation}
    \Delta_{v_j}F(U,V) = -2\sum_{i=1}^{n}u_i(r_{ij}-\left<u_i,v_j\right>)+2\lambda v_j
    \label{sgd4}
\end{equation}
The parameters are updated iteratively until the model loss function is less than a fixed threshold or the gradient difference between the two iterations is small, then the model can be considered as convergent.
\subsection{Distributed recommendation system}
We suppose that in a distributed environment, items are shared but different users belong to different data sources. For example, for users who buy iPhones, some users may purchase through Taobao, while others  will buy on the official website for quality reasons. 
Different users will rate the same item on different data sources. We assume that there are T data sources, the distributed matrix factorization recommendation system can be represented by Algorithm\ref{alg:DMF}.

\begin{algorithm}
    \caption{Distributed Matrix Factorization}
    \label{alg:DMF}
    \begin{algorithmic}
    \REQUIRE $U_1,U_2,..,U_T,V,\delta$
    \STATE data sources init their user profile matrix $U_t$
    \STATE server init item profile matrix $V,\delta$
    \REPEAT 
    \STATE \textbf{data sources update:}
    \FOR {$t=1;t<=T;t++$}
    \STATE $\Delta_{U_t}F(U,V)=-2(R_{U_t}-U_tV)V^T+2\mu U_t$
    \STATE $U_t^{new} = U_t^{old}-\alpha\Delta_{U_t}F(U,V)$
    \STATE $Graident_t= -2U^T(R_{U_t}-U_tV)+2\mu V$
    \STATE send $Graident_t$ to server
    \ENDFOR
    \STATE \textbf{server update:}
    \STATE receive $Graidents$ from data sources
    \STATE $G = \sum_{t=1}^{T}Graident_t$
    \STATE $V^{new} = V^{old} - G$
    \UNTIL{$G<\delta$}
    \end{algorithmic}
\end{algorithm}

Under this framework, each data source holds its user profile matrix and keeps it secret to the outside. 
the pubilc item profile matrix is stored in the central server. Each party uses the local rating matrix to update the user parameters, and only exposes the gradient of the item matrix to the server. 
The server updates the item matrix after summarizing the gradient. This method only involves the transmission of gradients, therefore the security of local data is protected. 
However,transmit gradient can also exposes privacy. knowing the gradients of a data source uploaded in two continuous steps, it can infer the rating information by the equations \eqref{leakage1}\eqref{leakage2}. And see more detail in the paper\cite{chai2019secure}. Therefore this article introduces secret sharing method in the transmission of graident, which make gradient transmission more efficient and safe.

\begin{equation}
    u_i^t=(r_{ij}-\left< u_i^t,v_j^t\right>)=G_j^t
    \label{leakage1}
\end{equation}

\begin{equation}
    r_{ij}=\frac{G_{jk}^t}{u_{ik}^t}+\sum_{m=1}^{D}u_{im}^t v_{jm}^t
    \label{leakage2}
\end{equation}
\subsection{Secret sharing}
The idea of secret sharing is to split the secret in an appropriate way, and each share after splitting is managed by different participants. A single participant cannot recover the secret information, and only several participants can cooperate to recover the secret message.

\begin{figure}[htbp]
    \centerline{\includegraphics[width=0.45\textwidth]{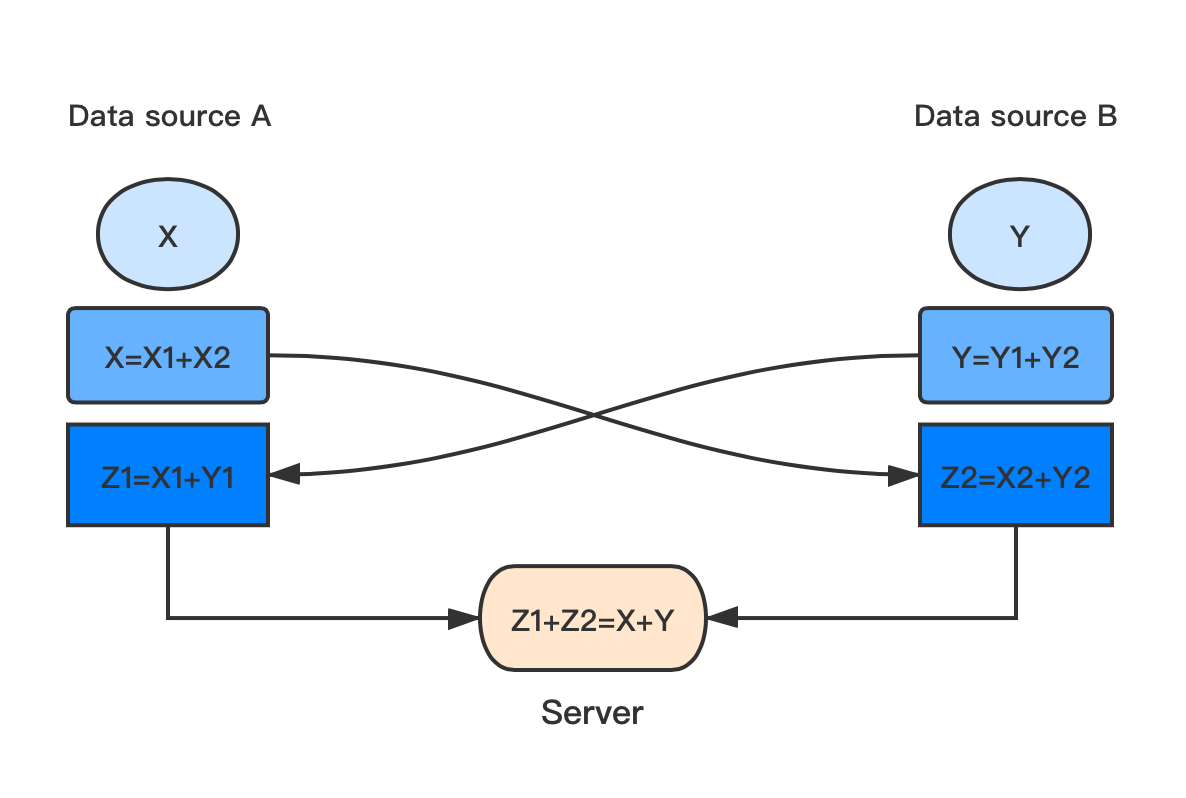}}
    \caption{An example of Secret sharing}
    \label{ss}
\end{figure}

The figure\ref{ss} gives a simple example of how to use sercet sharing. Two data sources own the number $X$ and $Y$ respectively, the server want to know the sum $X+Y$ but it will know nothing about X and Y. The process can be described as follows: 
firstly, the original data is decomposed into two sub parts, and one sub part is exchanged between the two sides, and then the sum of the remaining sub parts with the part from other side is calculated. Finally, the solution of the original problem is obtained by summarizing the calculated sum. 
In the process, the original data will not be exposed, so the sum operation can be completed under the premise of protecting data privacy. In addition, the multiplication can be realized by setting additional triples. In\cite{zheng2020industrial}, the author uses secret sharing technology to implement multi-source federated neural network.
\subsection{Put all together}
In order to solve the privacy problem that may be caused by the exposure gradient, we propose a shared matrix factorization (SMF) method based on secret sharing. 
As shown in the algorithm\ref{alg:SMF}, the data source calculates the local user profile matrix  parameters and the item matrix gradients are encrypted by secret sharing technology before transmitting to the server, and finally the encrypted gradients are summarized on the server to update the item profile matrix parameters.
\begin{algorithm}
    \caption{Shared Matrix Factorization}
    \label{alg:SMF}
    \begin{algorithmic}
    \REQUIRE $U_1,U_2,..,U_T,V,\delta$
    \STATE all parties initialize related parameters
    \REPEAT 
    \STATE \textbf{data sources update:}
    \FOR {$t=1;t<=T;t++$}
    \STATE update user profile matrix $U_t$
    \STATE compute item matrix gradient $g_t^{plain}$
    \STATE generate random number that meets $g_t^{plain}=g_{t}^{sub_1}+g_{t}^{sub_2}+..+g_{t}^{sub_T}$
    \STATE keep $g_t^{sub_t}$ and send the rest to other data 
    \STATE receive $g^{sub_t}$ from others 
    \STATE compute hybrid gradient $g_t^{hybrid}=\sum_{i=1}^T g_{i}^{sub_t}$
    \STATE send hybrid gradient to server
    \ENDFOR
    \STATE \textbf{server update:}
    \STATE receive $g^{hybrid}$ from data sources
    \STATE $G = \sum_{t=1}^{T}g_t^{hybrid}$
    \STATE $V^{new} = V^{old} - G$
    \UNTIL{$G<\delta$}
    \end{algorithmic}
\end{algorithm}

\section{Evaluation}
\subsection{Dataset}
To make the recommendation algorithm be better applied to the actual scene, we choose the real world dataset Movielens, which has been applied in many recommendation systems, such as caser\cite{tang2018personalized}, h4mf\cite{wang2018modeling}. 
We disorganize the rating matrix and randomly sampled the train/test set according to the ratio of 7:3.
\subsection{Parameters}
Through training experience and super parameter adjustment, we choose a group of better parameter combinations, in which the profile matrix dimension $k = 100$,
the regularization parameters $reg_u=10^{-3}, reg_v=10^{-3}$, and the learning rate is $lr=10^{-2}$
\subsection{Environment}
All experiments are performed on a server with 2.5GHz 16-core CPU and 64GB RAM, where the operation system is Linux and the program language is Python. 
We use multithreading to simulate multi-source data holder. And they communicate and exchange data through grpc. Each source will start a rpc server client to receive data from other clients 
\subsection{Performance}
\begin{figure}[htbp]
    \centerline{\includegraphics[width=0.45\textwidth]{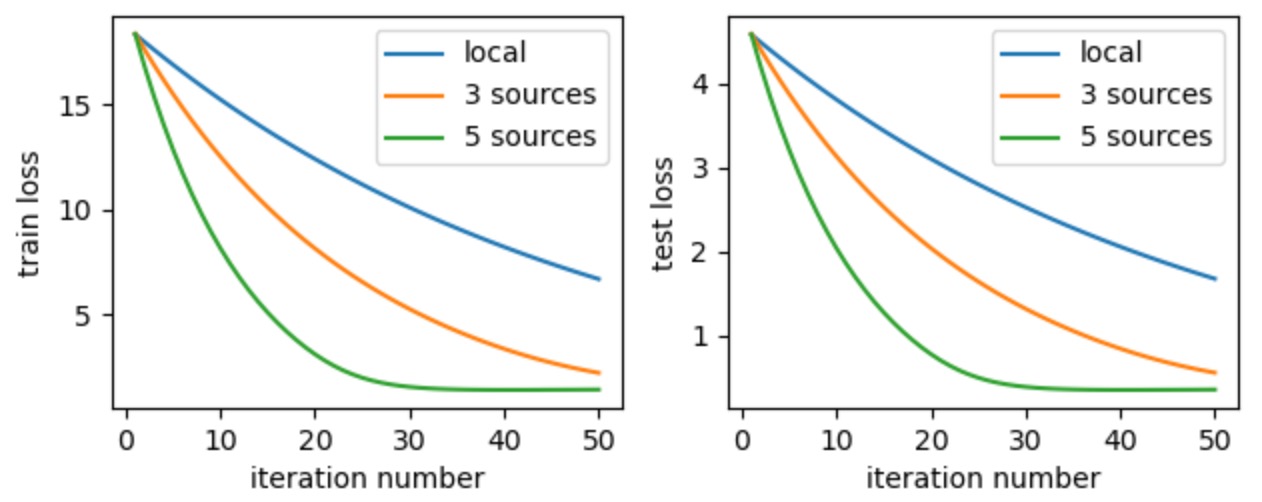}}
    \caption{local and distributed recommender system comparison}
    \label{ld}
\end{figure}

\paragraph{\textbf{local and distributed comparison}}
First, we tested the improvement that the distributed recommendation system can bring. We used the data provided by only local data, three data sources and five data sources. 
For each additional data source, the number of rating users increased by 200, and the total number of movies remained at 500. The experimental results are shown in the figure\ref{ld}. With the increase of data sources, the loss of the model decreases. 
This is due to the increase in the number of users, the rating matrix is more perfect, which makes the item vector fitting better.

\begin{figure}[htbp]
    \centerline{\includegraphics[width=0.40\textwidth]{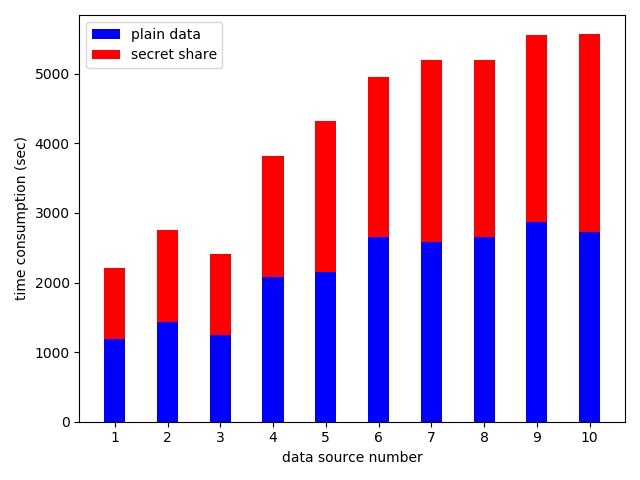}}
    \caption{time consumption with different data source numbers}
    \label{hc}
\end{figure}

\paragraph{\textbf{Horizontal comparison}}
We have tested the improvement brought by distributed recommendation. In federated learning, the main reason that affects the performance of distributed algorithms is the overhead of encryption methods. 
Therefore, we test the different performance between our algorithm and that without encryption. Since the main cost of secret sharing lies in the communication and exchange of sub secrets between nodes, we set different number of data sources for horizontal comparison. 
The result is as shown in the figure\ref{hc}. Compared with matrix factorization, the communication cost caused by secret sharing is less than the computation cost by matrix factorization. Therefore, the performance of shared MF is basically the same as that of common distributed recommendation system, which means our algorithm has strong practicability.

\begin{figure}[htbp]
    \centerline{\includegraphics[width=0.45\textwidth]{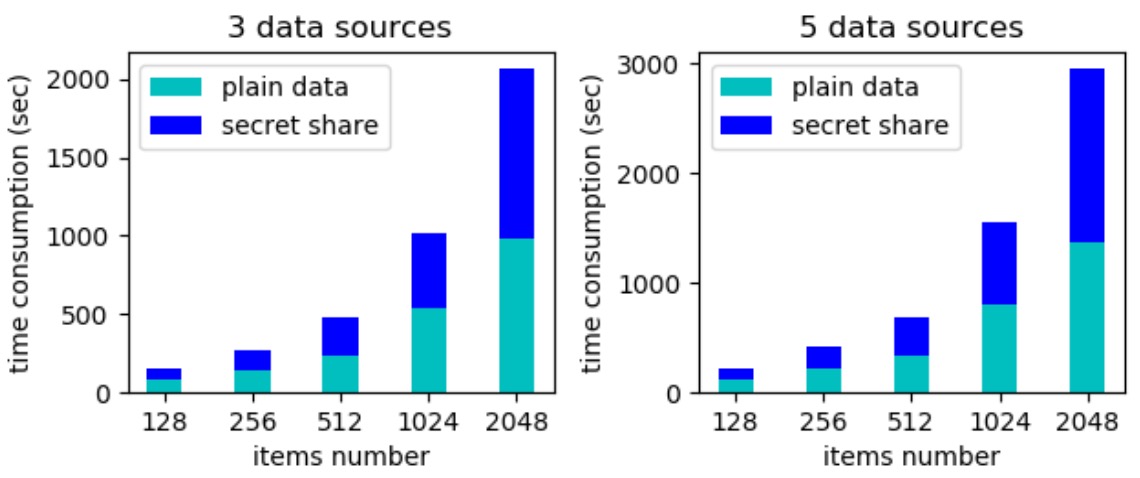}}
    \caption{time consumption with different item numbers}
    \label{vc}
\end{figure}

\paragraph{\textbf{Vertical comparison}}
In the previous horizontal comparison, we studied the communication overhead caused by increasing data sources. In the process of secret sharing of each data source, the amount of data transmitted is determined by the size of the item profile matrix. 
Therefore, we select the appropriate number of data sources and set different number of items to test the algorithm performance. The experimental results are shown in the figure\ref{vc}. There are three data sources on the left and five data sources on the right. 
It is obvious that with the increase of the number of objects, the communication overhead does not increase significantly, which proves  our algorithm is also very adaptable to large-scale items.

\paragraph{\textbf{why not homomorphic encryption}}
From the perspective of cryptography, homomorphic encryption can guarantee zero leakage of data privacy. Therefore, the distributed recommendation system using this method has the best security in theory. 

However, the disadvantage of homomorphic encryption is very obvious. The computational cost of data encryption and decryption process is very high. We compared our algorithm with FedML which uses an addition Encryption Paillier and tested the time cost under the same condition. 

From the table\ref{tab1}, we can see that homomorphic encryption scheme can work when the amount of data is small, but with the increase of data volume, the encryption time is obviously too high, which can not adapt to the actual large-scale recommendation scenarios.

\begin{table}[htbp]
    \caption{SharedMF vs FedML}
    \begin{center}
    \begin{tabular}{|c|c|c|c|}
    \hline
    train time(sec)&items50&items200 &items500  \\
    \hline
    FedML& 223.49 & 843.21 & 2064.62 \\
    \hline
    SharedMF& 100.58 & 284.13 & 583.37 \\
    \hline
    \end{tabular}
    \label{tab1}
    \end{center}
\end{table}

\section{Concluson and futrue work}
In this paper, we propose a secure distributed matrix factorization recommendation system framework, called SharedMF. Specifically, we first construct a distributed recommendation scenario, and store user data and item information separately in the clients and a server. 
The model is fitting by exchanging gradients between them, and the secret sharing technology is used to ensure the data privacy and security in the training process.

In the experimental stage, we first prove the usefulness of the distributed system to improve the accuracy of recommendation scenarios, and then compare the performance differences between our algorithm and the non-encrypted distributed recommendation to verify the practicability of the algorithm. 
Moreover, we test the existing solutions based on homomorphic encryption, which proves that our scheme is more robust to the increase in the number of users and items, and is more suitable for large-scale recommendation scenarios.

With the  importance of privacy protection in recommendation system and machine learning increasing, federated learning technology based on cryptography is bound to be widely used. 
The secret sharing technology used in this paper skilfully avoids the high computational complexity of traditional homomorphic encryption algorithm, and effectively improves the performance of privacy protection algorithm. However, it is worth mentioning that in this paper, secret sharing is only used to solve the privacy problem in the traditional algorithm matrix factorization. 
How to apply it in the current popular deep neural network will be our further research topic.

\bibliographystyle{mat/IEEEtran.bst}
\bibliography{mat/ref.bib}
\vspace{12pt}
\end{document}